\documentclass[sigconf,nonacm]{acmart}
\AtBeginDocument{%
  }
\usepackage{multirow,multicol}
\usepackage{fancyhdr}
\usepackage{booktabs}
\usepackage[table]{xcolor}
\usepackage{colortbl}
\usepackage{pifont}
\usepackage{makecell}
\usepackage{times}
\usepackage{epsfig}
\usepackage{graphicx}
\usepackage{amsmath}
\usepackage{algorithm}
\usepackage{algorithmic}
\usepackage{caption}
\usepackage{subcaption}

\renewcommand\footnotetextcopyrightpermission[1]{}

\definecolor{cleanrow}{RGB}{245, 245, 245}
\definecolor{mixedrow}{RGB}{255, 235, 235}
\definecolor{tttrow}{RGB}{230, 244, 255}
\definecolor{avgrow}{RGB}{245, 240, 255}
\definecolor{bestcolor}{RGB}{220, 20, 60}
\definecolor{baserow}{RGB}{245,245,245}      
\definecolor{singlerow}{RGB}{235,245,255}    
\definecolor{doublerow}{RGB}{255,245,230}    
\definecolor{fullrow}{RGB}{235,255,235}      

\definecolor{rainrow}{RGB}{240,248,255}    
\definecolor{fogrow}{RGB}{245,245,245}     
\definecolor{tttrow}{RGB}{235,255,235}     
\newcommand{\cmark}{\ding{51}}

\newcommand{\improve}[1]{\textcolor{blue}{\scriptsize($\downarrow$#1)}}
\newcommand{\best}[1]{\textbf{#1}}
\setcopyright{none}
\copyrightyear{2026}
\acmYear{2026}

\begin{document}

\title{Test-Time Training for Robust Text-Guided Open-Vocabulary Object Counting}


\author{Hao-Yuan Ma}
\affiliation{%
  \institution{School of Computer Science and Technology, Soochow University}
  \city{Suzhou}
  \country{China}}

\author{Yuda Zou}
\affiliation{%
  \institution{Wuhan University}
  \city{Wuhan}
  \country{China}}

\author{Li Zhang}
\affiliation{%
  \institution{School of Computer Science and Technology, Soochow University}
  \city{Suzhou}
  \country{China}}

\author{Yongchao Xu}
\affiliation{%
  \institution{Wuhan University}
  \city{Wuhan}
  \country{China}}

\renewcommand{\shortauthors}{Ma et al.}

\begin{abstract}
    Text-guided Open-vocabulary Object Counting (TOOC) enables counting arbitrary object categories specified by text prompts, offering substantially greater flexibility than conventional closed-set counting.
    However, existing TOOC methods are developed and evaluated primarily on ideal images, while real-world scenes often suffer from adverse conditions such as rain, fog, darkness, and sensor noise, which severely degrade visual quality and impair vision-language alignment. To bridge this gap, we introduce Robust-TOOC, the first benchmark for evaluating TOOC under diverse corruption conditions, which covers six representative degradation types: rain, fog, darkness, Gaussian noise, salt-and-pepper noise, and mixed corruption.
    To improve robustness while preserving the original counting architecture, we propose Dual-TTT, a dual-architecture test-time training framework for TOOC. Specifically, during test-time training, Dual-TTT updates only the Text-guided Lightweight Denoising module (TL-Denoiser), while keeping the original counting network frozen. Inspired by diffusion models, the TL-Denoiser is optimized to remove corruption-aware noise from image representations under degraded conditions. Since only the TL-Denoiser is trained at test time, Dual-TTT is annotation-free and can be seamlessly integrated into existing TOOC models without modifying their original architecture. Extensive experiments on multiple recent TOOC baselines demonstrate the effectiveness of our method. The code is provided in the supplementary material.
\end{abstract}



\begin{teaserfigure}
  \includegraphics[width=\textwidth]{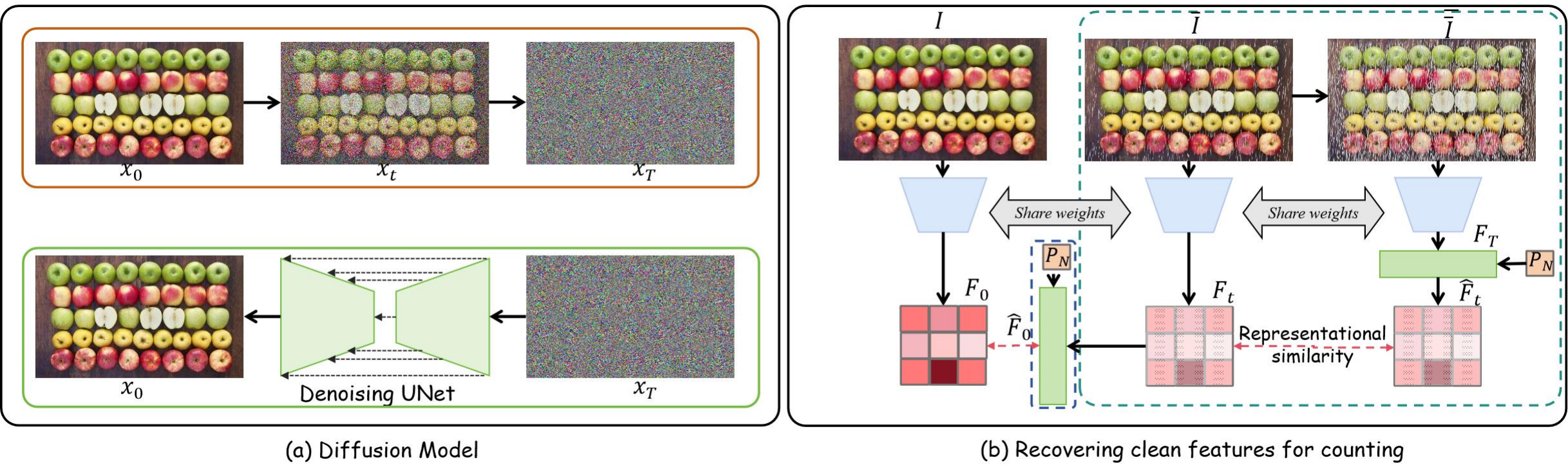}
  \caption{Motivation of Dual-TTT.
(a) Diffusion models reconstruct clean images from noisy inputs.
(b) Inspired by diffusion models, Dual-TTT performs feature-level denoising during test-time training. 
Specifically, it optimizes only the green module, the Text-guided Lightweight Denoising module (TL-Denoiser), to suppress corruption-induced noise in feature representations by maximizing the similarity between degraded features and their cleaner counterparts. During inference, the TOOC model uses the trained TL-Denoiser to reconstruct cleaner feature maps for robust counting.
}
  \label{fig:teaser_motivation}
\end{teaserfigure}


\maketitle

\section{Introduction}
Object counting is a fundamental vision problem with wide applications in traffic analysis \cite{TRANCOS}, retail intelligence \cite{warehouse}, and UAV inspection \cite{CarPK}. 
Most existing counting methods \cite{FGENet,P2PNet,vmambacc,M2PLNet} are developed in a closed-set setting, where target object categories are predefined during training. However, closed-set methods lack the flexibility needed in real-world applications, as the model usually has to be retrained whenever users want to count a new object category. 
Recently, Text-guided Open-Vocabulary Object Counting (TOOC) has emerged as a promising direction by enabling category specification through text prompts, thereby breaking the category boundary of conventional counting models.

Despite recent progress, existing TOOC methods are almost exclusively studied under ideal and clean image conditions. In practice, images acquired in real-world environments are often affected by adverse weather and sensor corruptions, such as rain, fog, darkness, Gaussian noise, and salt-and-pepper noise. Under these degradations, the performance of current TOOC models \cite{CLIP-Count,OVID-ma,CountGD} drops drastically. 
Unlike conventional closed-set counting, TOOC methods are heavily based on visual-language alignment to count target objects. When corruptions distort category discriminative visual patterns, the alignment between image features and textual queries becomes unreliable \cite{clipnoise}, resulting in severe counting errors.
Therefore, robust text-guided open-vocabulary object counting constitutes an important yet largely unexplored problem.

Improving robustness through corruption-aware training is a natural direction. However, this strategy is often impractical for TOOC, since real-world degradations are diverse, unknown, and difficult to enumerate in advance.
Moreover, retraining large vision-language counting models for every corruption setting is computationally expensive and undermines deployment flexibility. 
This motivates us to explore test-time training \cite{ttt_ICML}, which adapts the model during inference using only test samples without requiring manual annotations.
Nevertheless, existing test-time training or adaptation methods are mostly designed for classification \cite{clipnoise} or detection \cite{ioufilter}, and generally perform category-agnostic feature stabilization.
This design is suboptimal for TOOC, where the adaptation process should be explicitly guided by the text prompt and focus on the target category of interest.

To address the above challenges, we investigate robust TOOC from both benchmark and framework perspectives. First, we introduce Robust-TOOC, a comprehensive benchmark for evaluating TOOC models under six representative degradation types, including rain, fog, darkness, gaussian noise, salt-and-pepper noise, and mixed corruption. 
Second, to improve the robustness of TOOC methods, we propose Dual-TTT, a novel text-guided test-time training framework. Specifically, Dual-TTT adopts a dual-branch architecture, where the degraded input and its dynamically re-noised counterpart are jointly processed to learn spatially invariant counting cues during test-time training. This training process is explicitly guided by the text prompt, making it prompt-aware and category-specific. As shown in Fig~\ref{fig:teaser_motivation}, we design a plug-and-play denoising module within this framework, termed TL-Denoiser, inspired by the denoising principle of diffusion models \cite{diffusion_org,ddim,ddpm}, to restore corrupted feature representations at test time. TL-Denoiser explicitly recovers semantically aligned visual tokens under text guidance, and it is the only component optimized during test-time training, enabling seamless adaptation of existing TOOC architectures without requiring additional annotations.

Our contributions are summarized as follows:
\begin{itemize}

    \item We build the first systematic investigation of robust TOOC, and establish Robust-TOOC, a new benchmark covering six diverse real-world degradation types.

    \item We propose the Dual-TTT framework, the first framework that explicitly incorporates text guidance into test-time training for TOOC, enabling prompt-aware adaptation toward the target category without requiring extra annotations. 

    \item Extensive experiments on multiple TOOC baselines demonstrate that Dual-TTT achieves significant and consistent robustness improvements across diverse types and severity levels of degradation.

\end{itemize}
    
\section{Related Work}

\subsection{Text-Guided Open-Vocabulary Object Counting}
Recent text-guided open-vocabulary object counting (TOOC) methods extend class-specific counting by conditioning the model on free-form category text rather than fixed label sets. Early CLIP-based pipelines (e.g., CLIP-Count \cite{CLIP-Count}, CounTX \cite{CounTX}) mainly follow a density-regression paradigm: they align patch-level visual features with text prompts and decode density maps for counting. These methods show strong zero-shot transfer, but are mostly developed and validated on clean images.

Another line of research reuses open-vocabulary detection or vision-language foundation models for counting. CountGD \cite{CountGD} shows that grounding-style localization can significantly improve open-world counting accuracy. DCount \cite{DCount} further exploits detection-oriented visual-text representations to enhance category-aware counting, while SDVPT \cite{zhao2025sdvpt} investigates prompt-based transfer to adapt pretrained vision-language models to counting scenarios. TrueCount \cite{truecount} further improves counting performance by introducing depth cues \cite{yang2024depth} and SAM-based priors to provide richer structural and object-aware information. More recently, LGCount \cite{LGCount} emphasizes that counting involves not only local visual matching but also global number reasoning, and thus introduces local ranking and number-aware prompt alignment. However, these methods still largely assume benign visual conditions, and robustness to realistic corruptions, including rain, fog, low light, and sensor noise, remains underexplored.

\subsection{Test-Time Adaptation and Test-Time Training}
Test-time Training/Adaptation (TTT/TTA) updates model parameters using unlabeled test data to handle distribution shifts \cite{ttt_ICML}. In detection and related dense prediction tasks, existing strategies include self-training with pseudo labels, feature-distribution alignment, one-shot adaptation, and continual test-time adaptation with memory/skip-update mechanisms \cite{test,ioufilter,tttp}. These methods consistently show that adaptation can recover performance under domain shift, especially when source data is unavailable.

Despite this progress, most existing TTT/TTA frameworks are designed for unimodal outputs, such as classification and detection, and mainly optimize generic feature-stability objectives. They neither explicitly exploit user-provided text prompts to guide adaptation toward category-relevant counting cues, nor directly address cross-modal token misalignment caused by image corruptions. In contrast, robust TOOC requires both resilience to degradation and prompt-conditioned semantic consistency for accurate counting.

Therefore, we establish Robust-TOOC, a benchmark for evaluating TOOC under diverse visual corruptions, and propose Dual-TTT, a text-guided test-time training framework for robust open-vocabulary counting. Specifically, our method combines a degraded/re-noised dual-branch interaction strategy for learning degradation-invariant spatial representations with a text-guided token denoising module that restores visual-text semantic alignment under corruptions.
\begin{figure}[t]
    \centering
    \includegraphics[width=1.0\linewidth]{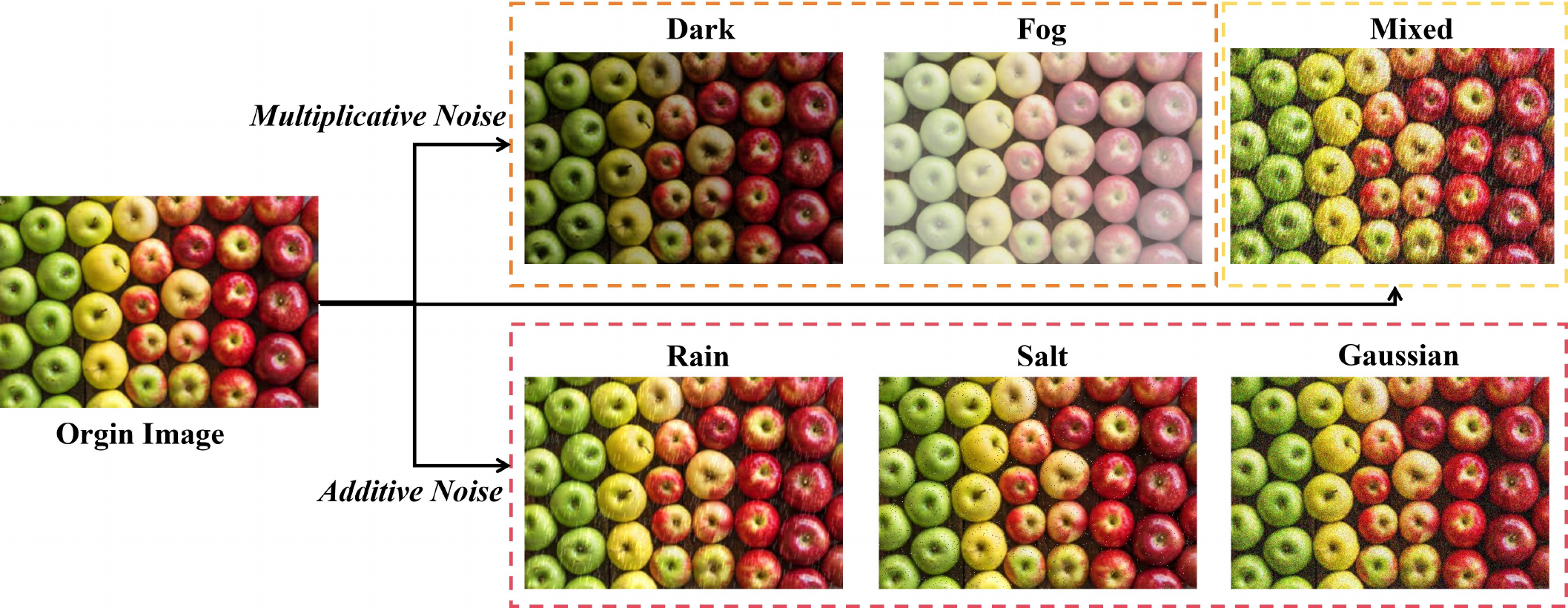}
    \caption{Examples of image degradations in Robust-TOOC. From a clean image, we generate six representative corruption types—rain, fog, darkness, gaussian noise, salt-and-pepper noise, and mixed corruption—to evaluate robust text-guided open-vocabulary object counting in real-world degraded scenarios.}
    \label{fig:noise_type}
\end{figure}
\section{Robust-TOOC Benchmark}

\subsection{Benchmark Construction}
To systematically evaluate robustness in text-guided open-vocabulary object counting, we construct \textbf{Robust-TOOC} by corrupting images from the original clean TOOC benchmark (FSC147 \cite{FamNet}) while keeping all text prompts and point annotations unchanged. This design isolates the impact of visual degradation on counting performance and alignment in the vision-language, without introducing additional annotation bias.

Formally, given an image-prompt pair $\langle p, I \rangle$ with count label $y$, we generate a corrupted image $\tilde{I}^{(c)}$ under corruption type $c$ and obtain the benchmark tuples:
\begin{align}
\mathcal{D}_{\text{robust}}=\{(\tilde{I}^{(c)}, p, y)\}.
\end{align}

We apply corruptions only to the validation and test splits while keeping the training split clean. This protocol follows a realistic deployment setting, where models trained on clean source data are evaluated under unseen target-domain degradations, and it avoids contaminating the source-domain training distribution.

We intentionally construct Robust-TOOC directly in the image space, rather than relying on reconstruction-based generation pipelines such as diffusion models \cite{ddim,ddpm}. Although diffusion-based corruption synthesis or restoration can produce visually plausible results, such methods are often unreliable in dense counting scenarios. In particular, the VAE-based \cite{vae} encoding-decoding process may alter fine-grained spatial details, causing annotation misalignment, local structure distortion, or even disappearance of small target instances. These effects are especially problematic for counting benchmarks, where accurate object localization and point-level correspondence are crucial. By preserving the original annotations and introducing only controlled image degradations, our benchmark provides a more faithful and reproducible testbed for robust TOOC evaluation.
All corruption generation is deterministic under fixed random seeds to ensure reproducibility.

\begin{figure*}
    \centering
    \includegraphics[width=0.8\linewidth]{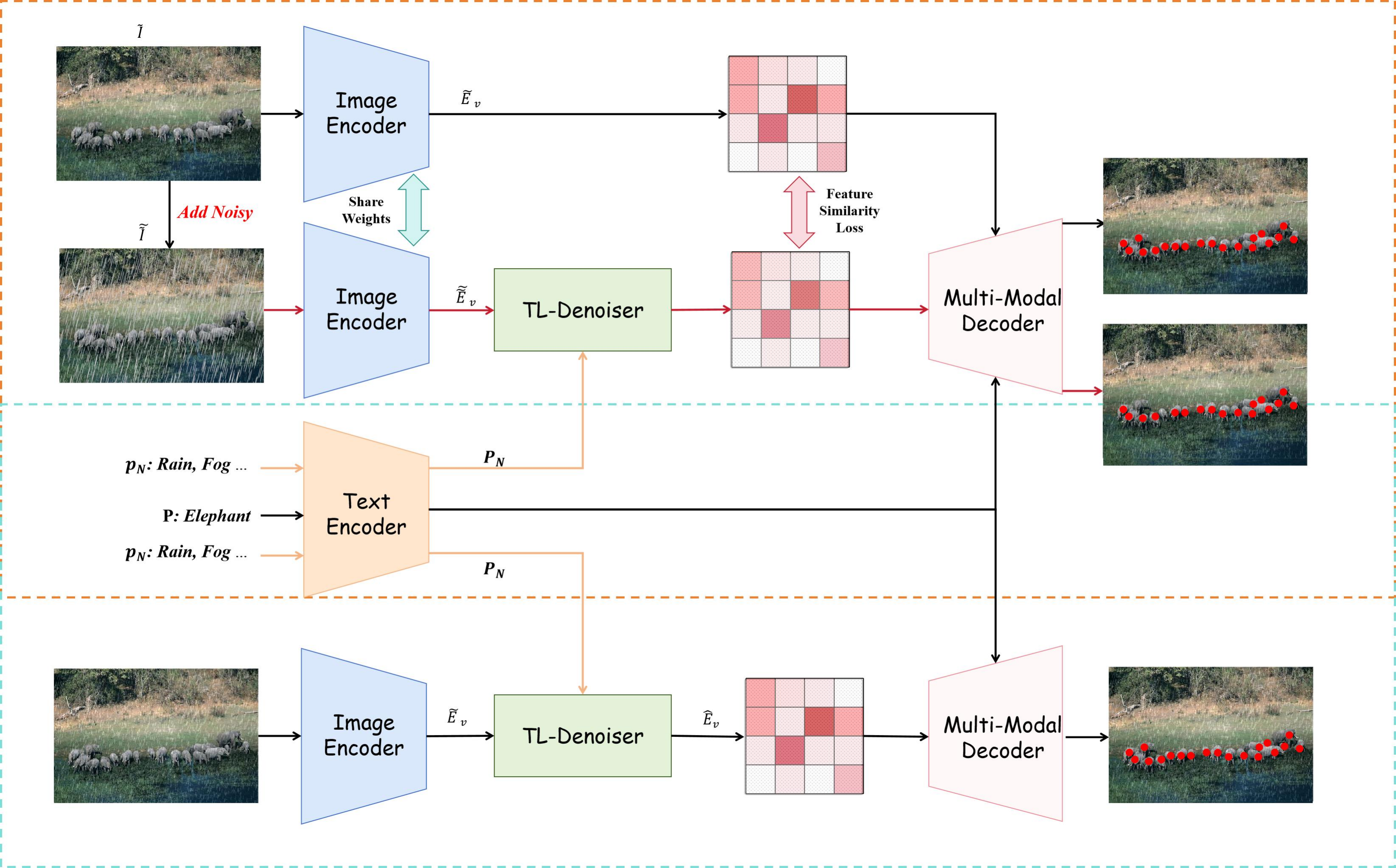}
    \caption{Overview of Dual-TTT.
The orange dashed box denotes the test-time training stage, while the cyan dashed box denotes the inference stage. During test-time training, the corrupted input and its artificially re-noised counterpart are fed into two weight-sharing image encoders. The red arrows indicate the flow of corrupted-image features, which are further processed by the TL-Denoiser to suppress corruption-induced noise in the representation space. Meanwhile, the orange arrows denote the flow of the corruption prompt $P_N$, which is encoded by the text encoder and used to guide the denoising process. A feature similarity loss is imposed between the cleaner branch and the denoised branch to encourage corruption-invariant representations. During inference, only the lower branch is retained: the corrupted image is encoded, refined by the TL-Denoiser, and then sent to the multi-modal decoder for robust text-guided counting.}
    \label{fig:ttt_dual}
\end{figure*}

\subsection{Corruption Types and Severity Settings}
As shown in Fig.~\ref{fig:noise_type}, the Robust-TOOC covers six representative corruption types: \textbf{rain}, \textbf{fog}, \textbf{darkness}, \textbf{Gaussian noise}, \textbf{salt-and-pepper noise}, and \textbf{mixed corruption}. These corruptions span weather-induced degradation, illumination variation, and sensor noise commonly encountered in real-world applications.

To improve physical plausibility, we distinguish between indoor and outdoor scenes before applying weather effects. Specifically, we use GPT-4o together with human verification to determine scene type, and apply rain or fog only to images with appropriate outdoor content. In this way, the generated corruptions better reflect realistic environmental conditions.

For physical realism, fog is synthesized with a depth-aware atmospheric scattering model:
\begin{align}
I(x)=J(x)\,t(x)+A(1-t(x)), \quad t(x)=\exp(-\beta d(x)),
\end{align}
where $d(x)$ is the estimated depth obtained by Depth Anything V2 \cite{yang2024depth}, $\beta$ controls the scattering strength, and $A$ denotes atmospheric light. Other corruptions are generated using parameterized stochastic processes, such as the density, length, and angle of the rain streak, the attenuation of low-light, the Gaussian noise level and the impulse noise ratio.

For mixed corruption, we randomly composed two or three basic corruption types to simulate more complex real-world degradations. This setting enables a comprehensive evaluation of the robustness of the model under both single-factor and compound corruptions.


\section{Method}

\subsection{The Dual-TTT Framework}
\label{sec:overview_tooc}

Dual-TTT is a test-time training framework designed for robust text-guided counting under real-world degradations.

Under ideal conditions, given a text-image pair $\langle p, I \rangle$, text-guided open-vocabulary object counting (TOOC) mainly relies on the cross-modal alignment learned by a pre-trained vision-language model (VLM). Specifically, the text prompt $p$ is first encoded by a text encoder to obtain the textual embedding
\begin{equation}
E_t = \psi_t(p),
\end{equation}
where $\psi_t(\cdot)$ denotes the text encoder. Meanwhile, the input image $I$ is processed by an image encoder to extract the visual embedding
\begin{equation}
E_v = \psi_v(I),
\end{equation}
where $\psi_v(\cdot)$ denotes the image encoder.

The visual and textual embeddings are then fed into a multi-modal decoder for cross-modal reasoning. In general, the decoder consists of a Transformer and a Cross-Attention module. The Transformer performs global contextual modeling over $E_v$, while the Cross-Attention module uses the textual feature as the key and value, and the updated visual feature as the query:
\begin{align}
E_v^i &= \mathrm{MHSA}(E_v^{i-1}), \\
Q &= W_Q E_v^i, \qquad K = W_K E_t, \qquad V = W_V E_t, \\
\mathrm{Attn}(Q, K, V) &= \mathrm{softmax}\left(\frac{QK^\top}{\sqrt{d}}\right)V,
\end{align}
where $i$ denotes the decoder block index, $\mathrm{MHSA}$ is multi-head self-attention, and $d$ is the feature dimension.

The decoder output is further fed into a prediction head to estimate either a density map $D$ or a set of object coordinates. Accordingly, the final count $\hat{y}$ can be obtained as:
\begin{equation}
D = f_{\mathrm{head}}(E_v^l), \qquad
\hat{y} = \sum_{i,j} D_{i,j},
\end{equation}
or
\begin{equation}
\mathcal{C} = f_{\mathrm{head}}(E_v^l), \qquad
\hat{y} = |\mathcal{C}|,
\end{equation}
where $E_v^l$ denotes the visual embedding from the last decoder block and $\mathcal{C}$ denotes the predicted coordinate set.

However, in practical scenarios, the input image is often degraded by adverse factors such as noise, blur, illumination variation, or weather corruption. In this case, the VLM receives a corrupted pair $\langle p, \tilde{I} \rangle$, and the extracted visual feature becomes
\begin{equation}
\tilde{E}_v = \psi_v(\tilde{I}), \qquad \tilde{E}_v \in \mathbb{R}^{N \times C},
\end{equation}
where $N$ is the number of visual tokens and $C$ is the channel dimension. Our goal is to recover a clean-equivalent representation from $\tilde{E}_v$ using the proposed TL-Denoiser:
\begin{equation}
\hat{E}_v = \mathrm{TL\text{-}Denoiser}(\tilde{E}_v, P_N)
\end{equation}
\begin{equation}
\hat{E}_v \approx E_v
\end{equation}

Inspired by diffusion models~\cite{diffusion_org,ddpm}, which learn to predict the clean image $I$ from a noisy observation $I^t$ at time step $t$, we reformulate the restoration problem in the \emph{feature space} rather than the pixel space. Instead of relying on a predefined forward diffusion process, we construct a self-supervised denoising objective during test time. 
Different from diffusion models, where the denoising process is conditioned on the time step $t$, we condition our restoration on \emph{noise semantics}. Specifically, we use the text encoder $\psi_t$ to encode a set of noise-type prompts into text prototypes
\begin{equation}
P_N = \psi_t(p_N),
\end{equation}
where $p_N$ denotes the textual descriptions of different degradation types. 

As shown in Fig.~\ref{fig:ttt_dual}, given a degraded image $\tilde{I}$, we further apply an additional perturbation operator $\mathcal{A}(\cdot)$ to obtain a more severely corrupted view:
\begin{equation}
\tilde{\tilde{I}} = \mathcal{A}(\tilde{I}).
\end{equation}
Feeding the two views into the image encoder:
\begin{equation}
\tilde{E}_v = \psi_v(\tilde{I}), \qquad
\tilde{\tilde{E}}_v = \psi_v(\tilde{\tilde{I}}).
\end{equation}
We then train TL-Denoiser at test time to map the more degraded feature $\tilde{\tilde{E}}_v$ toward the less degraded feature $\tilde{E}_v$:
\begin{equation}
\mathrm{TL\text{-}Denoiser}(\tilde{\tilde{E}}_v, P_N) \approx \tilde{E}_v.
\end{equation}
After such online adaptation, the learned TL-Denoiser is used during inference to transform $\tilde{E}_v$ into a cleaner feature that approximates the latent clean representation $E_v$:
\begin{equation}
\mathrm{TL\text{-}Denoiser}(\tilde{E}_v, P_N) \approx E_v.
\end{equation}
Benefiting from the cross-modal alignment of the pre-trained VLM, we compute the similarity and entropy between each visual patch and the noise prototypes in $P_N$, thereby obtaining a patch-wise noise distribution. The entropy of this distribution reflects the uncertainty of degradation for each patch and serves as an adaptive denoising condition, playing a role analogous to the diffusion time step $t$.

\begin{figure}
    \centering
    \includegraphics[width=0.9\linewidth]{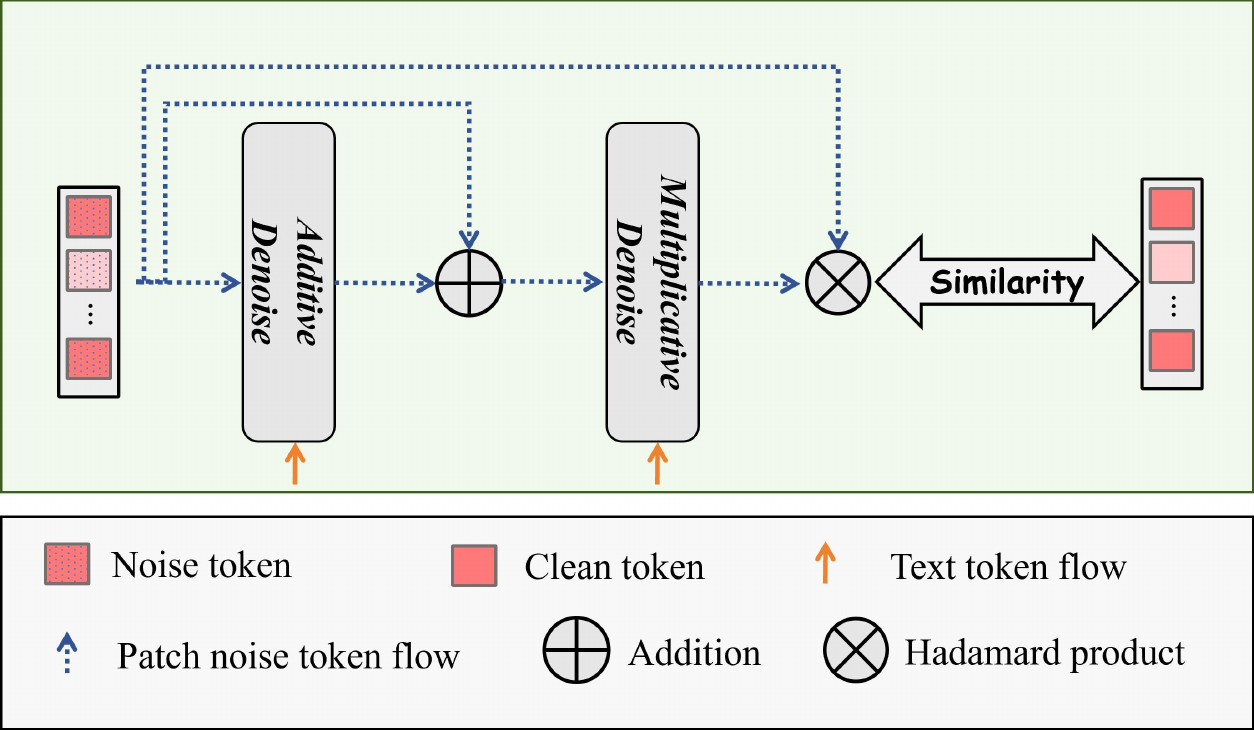}
    \caption{Illustration of the proposed TL-Denoiser. Noise-aware tokens are first processed by an additive denoising branch and then refined by a multiplicative denoising branch. The restored visual tokens are further aligned with clean tokens through a similarity-guided operation. The module leverages both patch-level noise token flow and text-token guidance to adaptively recover degradation-corrupted features.}
    \label{fig:ttt_denoise}
\end{figure}

\subsection{TL-Denoiser}

In the denoising process, visual features from different spatial regions of images are usually affected by degradations to different extents. Therefore, applying the same restoration strength to all locations is suboptimal, as it cannot properly distinguish target-related regions from irrelevant background areas. To address this issue, we exploit the alignment capability of the pre-trained VLM between image patches and text prompts to guide region-aware denoising, such that the restoration process can focus more on target-relevant regions.

To better characterize real degradations, we decompose image corruption into two broad categories: additive noise and multiplicative noise. Accordingly, we model the degradation process in feature space as:
\begin{equation}
\tilde{E}_v = \mathcal{N}_m \otimes E_v + \mathcal{N}_a,
\end{equation}
where $\mathcal{N}_m$ denotes the multiplicative degradation component, $\mathcal{N}_a$ denotes the additive degradation component, and $\otimes$ represents element-wise multiplication. Restoring the clean visual representation therefore requires estimating both types of corruption:
\begin{equation}
E_v = \mathcal{D}_m \otimes \tilde{E}_v + \mathcal{D}_a,
\end{equation}
where $\mathcal{D}_m$ and $\mathcal{D}_a$ denote the learned multiplicative and additive denoising terms, respectively.

Specifically, we first encode the degradation type into a noise prompt embedding $P_N \in \mathbb{R}^{N_t \times C}$,
where $N_t$ denotes the number of noise prompt tokens and $C$ is the feature dimension. We then compute the cosine similarity between $P_N$ and the degraded visual feature $\bar{E}_v$, yielding a noise-aware similarity matrix:
\begin{equation}
S_{\mathcal N} = \mathrm{CosSim}(p_{\mathcal N}, \bar{E}_v), \qquad S_{\mathcal N} \in \mathbb{R}^{N_t \times N},
\end{equation}
where $N$ is the number of visual tokens. This matrix characterizes the correlation between each visual location and the degradation pattern, thereby reflecting the degree to which different regions are affected by the current corruption.

As shown in Fig.~\ref{fig:ttt_denoise}, given $S_{\mathcal N}$ and $\tilde{E}_v$, we employ a lightweight prediction network to estimate a location-aware multiplicative restoration factor $\hat{\mathcal D}_{\mathcal N}$ and additive compensation term $\hat{b}_{\mathcal D}$:

\begin{equation}
\hat{\mathcal D}_{\mathcal N}
= g_{{\theta}_m}(S_{\mathcal N}, \bar{E}_v),\quad
 \hat{b}_{\mathcal D}
= g_{{\theta}_a}(S_{\mathcal N}, \bar{E}_v),
\end{equation}
where $g_{{\theta}_m}(\cdot)$ and $g_{{\theta}_a}(\cdot)$ denotes the lightweight restoration network. The preliminarily restored feature is then obtained as:
\begin{equation}
\hat{E}_v = \hat{\mathcal D}_{\mathcal N} \otimes \bar{E}_v + \hat{b}_{\mathcal D}.
\end{equation}

Furthermore, we use the entropy of the similarity distribution at each spatial location to determine whether restoration should be strengthened and by how much. For the $i$-th visual token, we first normalize its similarity scores over all noise prompt tokens:
\begin{equation}
\tilde{S}_{\mathcal N}^{(i)} = \mathrm{softmax}(S_{\mathcal N}^{(:,i)}),
\end{equation}
and compute the corresponding entropy as
\begin{equation}
H_i = - \sum_{j=1}^{N_t} \tilde{S}_{\mathcal N}^{(j,i)} \log \tilde{S}_{\mathcal N}^{(j,i)}.
\end{equation}
A lower entropy indicates that the current location is more confidently associated with a specific degradation pattern, and thus stronger restoration should be applied. In contrast, a higher entropy suggests a more ambiguous degradation pattern, for which a more conservative adjustment is preferred.

Accordingly, we generate an entropy-guided modulation weight
\begin{equation}
\alpha_i = 1 - \frac{H_i}{\log N_t},
\end{equation}
and use it to modulate the restoration parameters:
\begin{equation}
\mathcal D_{\mathcal N}^{(:,i)} = \alpha_i \hat{\mathcal D}_{\mathcal N}^{(:,i)} + (1-\alpha_i),
\end{equation}
\begin{equation}
b_{\mathcal D}^{(:,i)} = \alpha_i \hat{b}_{\mathcal D}^{(:,i)}.
\end{equation}

Finally, the refined denoised feature is formulated as:
\begin{equation}
\hat{E}_v = \mathcal D_{\mathcal N} \otimes \tilde{E}_v + b_{\mathcal D},
\end{equation}
which is expected to approach the clean visual representation:
\begin{equation}
\hat{E}_v \rightarrow E_v.
\end{equation}


\begin{table*}[t]
\centering
\small
\setlength{\tabcolsep}{6pt}
\renewcommand{\arraystretch}{1.15}
\caption{Performance comparison under clean and corrupted settings.}
\label{tab:main_results}
\begin{tabular}{ll l cccc}
\toprule
\multirow{2}{*}{Method} & \multirow{2}{*}{Model} & \multirow{2}{*}{Type} 
& \multicolumn{2}{c}{Val} & \multicolumn{2}{c}{Test} \\
\cmidrule(lr){4-5} \cmidrule(lr){6-7}
& & & MAE $\downarrow$ & RMSE $\downarrow$ & MAE $\downarrow$ & RMSE $\downarrow$ \\
\midrule

\multirow{5}{*}{CounTX}
& \multirow{5}{*}{OpenCLIP}
& \cellcolor{cleanrow} Clean
& \cellcolor{cleanrow} 17.11
& \cellcolor{cleanrow} 65.60
& \cellcolor{cleanrow} 15.88
& \cellcolor{cleanrow} 106.29 \\
&
& \cellcolor{mixedrow} Mixed
& \cellcolor{mixedrow} 42.34
& \cellcolor{mixedrow} 126.28
& \cellcolor{mixedrow} 25.30
& \cellcolor{mixedrow} 117.06 \\
&
& \cellcolor{tttrow} Ours Mixed
& \cellcolor{tttrow} \best{34.50}\,\improve{18.52\%}
& \cellcolor{tttrow} \best{102.23}\,\improve{19.05\%}
& \cellcolor{tttrow} \best{21.72}\,\improve{14.15\%}
& \cellcolor{tttrow} \best{110.96}\,\improve{5.21\%} \\
&
& \cellcolor{avgrow} Average
& \cellcolor{avgrow} 21.24
& \cellcolor{avgrow} 82.31
& \cellcolor{avgrow} 17.22
& \cellcolor{avgrow} 108.25 \\
&
& \cellcolor{tttrow} Ours Average
& \cellcolor{tttrow} \best{20.53}\,\improve{3.34\%}
& \cellcolor{tttrow} \best{76.93}\,\improve{6.54\%}
& \cellcolor{tttrow} \best{16.73}\,\improve{2.85\%}
& \cellcolor{tttrow} \best{108.06}\,\improve{0.18\%} \\
\midrule

\multirow{5}{*}{CLIP-Count}
& \multirow{5}{*}{CLIP}
& \cellcolor{cleanrow} Clean
& \cellcolor{cleanrow} 18.89
& \cellcolor{cleanrow} 67.57
& \cellcolor{cleanrow} 17.65
& \cellcolor{cleanrow} 109.01 \\
&
& \cellcolor{mixedrow} Mixed
& \cellcolor{mixedrow} 45.35
& \cellcolor{mixedrow} 139.29
& \cellcolor{mixedrow} 25.85
& \cellcolor{mixedrow} 112.66 \\
&
& \cellcolor{tttrow} Ours Mixed
& \cellcolor{tttrow} \best{28.32}\,\improve{37.55\%}
& \cellcolor{tttrow} \best{94.34}\,\improve{32.27\%}
& \cellcolor{tttrow} \best{21.72}\,\improve{15.98\%}
& \cellcolor{tttrow} \best{110.62}\,\improve{1.81\%} \\
&
& \cellcolor{avgrow} Average
& \cellcolor{avgrow} 24.90
& \cellcolor{avgrow} 84.17
& \cellcolor{avgrow} 19.49
& \cellcolor{avgrow} 109.77 \\
&
& \cellcolor{tttrow} Ours Average
& \cellcolor{tttrow} \best{20.80}\,\improve{16.47\%}
& \cellcolor{tttrow} \best{76.02}\,\improve{9.68\%}
& \cellcolor{tttrow} \best{18.32}\,\improve{6.00\%}
& \cellcolor{tttrow} \best{109.23}\,\improve{0.49\%} \\
\midrule

\multirow{5}{*}{CountGD}
& \multirow{5}{*}{GDNIO}
& \cellcolor{cleanrow} Clean
& \cellcolor{cleanrow} 11.19
& \cellcolor{cleanrow} 56.57
& \cellcolor{cleanrow} 13.32
& \cellcolor{cleanrow} 133.33 \\
&
& \cellcolor{mixedrow} Mixed
& \cellcolor{mixedrow} 17.24
& \cellcolor{mixedrow} 83.13
& \cellcolor{mixedrow} 16.60
& \cellcolor{mixedrow} 135.41 \\
&
& \cellcolor{tttrow} Ours Mixed
& \cellcolor{tttrow} 13.40\,\improve{22.27\%}
& \cellcolor{tttrow} 65.76\,\improve{20.90\%}
& \cellcolor{tttrow} 14.57\,\improve{12.23\%}
& \cellcolor{tttrow} 133.95\,\improve{1.08\%} \\
&
& \cellcolor{avgrow} Average
& \cellcolor{avgrow} 13.10
& \cellcolor{avgrow} 65.87
& \cellcolor{avgrow} 14.49
& \cellcolor{avgrow} 134.32 \\
&
& \cellcolor{tttrow} Ours Average
& \cellcolor{tttrow} 11.17\,\improve{14.73\%}
& \cellcolor{tttrow} 55.90\,\improve{15.14\%}
& \cellcolor{tttrow} 13.35\,\improve{7.87\%}
& \cellcolor{tttrow} 133.39\,\improve{0.69\%} \\
\midrule

\bottomrule
\end{tabular}
\end{table*}

\subsection{Training and Inference}

Our method performs online adaptation at test time without requiring any additional annotations. The core objective is to make the restored visual representation of a degraded image as close as possible to the clean visual representation, thereby maintaining stable counting performance under corruptions.

Specifically, given the degraded visual feature $\tilde{E_v}$ and its restored counterpart $\hat{\tilde{E}}_v$, we measure their discrepancy using the Kullback--Leibler (KL) divergence. Since raw visual features are not valid probability distributions, we first apply a softmax normalization along the token (channel) axis to convert each feature into a distribution before computing the divergence:
\begin{equation}
P = \mathrm{softmax}(\hat{\tilde{E}}_v), \qquad
Q = \mathrm{softmax}(\tilde{E_v}),
\end{equation}
where the softmax is taken over the feature dimension so that $\sum_{i} P^{(i)} = \sum_{i} Q^{(i)} = 1$. The test-time optimization objective is then defined as
\begin{equation}
\mathcal{L}_{\mathrm{KL}} 
= D_{\mathrm{KL}}\big( P \,\|\, Q \big)
= \sum_{i} P^{(i)} \log \frac{P^{(i)}}{Q^{(i)}}.
\end{equation}
Compared with directly matching feature magnitudes (e.g., MSE), this distributional formulation focuses on the \emph{relative} response pattern across tokens, which is more consistent with the counting objective. The same softmax normalization is applied identically during both test-time training and inference, ensuring that the restored and reference features are always compared as distributions.
By minimizing this objective, the proposed TL-Denoiser progressively corrects the representation shift caused by degradations in an annotation-free manner, so that the restored feature can be better aligned with the clean one.

In practical inference, to avoid additional computational overhead, we only keep a single image encoder. The test image is first processed by the image encoder to obtain degraded visual features, which are then restored by the TL-Denoiser. The adapted features are finally fed into the multi-modal decoder for object counting prediction. In this way, our framework can be seamlessly integrated into existing TOOC models without retraining the backbone or requiring extra supervision.

\section{Experiments}

\subsection{Experimental Settings}
Experimental Details.
All experiments are conducted under the PyTorch 2.0.1 framework with CUDA 11.6, using an NVIDIA RTX 3090 GPU with 24GB memory. To ensure reproducibility, the random seed is fixed to 42. We adopt Adam as the optimizer, with the learning rate set to $5 \times 10^{-4}$, weight decay set to $1 \times 10^{-4}$, and the similarity threshold set to 0.3. We compare our method against five representative TOOC baselines, including CounTX, CLIP-Count and CountGD. Following prior work, we use mean absolute error (MAE) and root mean square error (RMSE) as the evaluation metrics.

\subsection{Main Results}
Table~\ref{tab:main_results} shows that our method consistently improves performance over both the Mixed and Average baselines (mean performance over rain, fog, darkness, Gaussian noise, and salt-and-pepper noise) across different counting models and evaluation settings. 

For the CLIP-based models, the gains are especially clear. On CounTX, our method reduces the error over the Mixed baseline by 18.52\% / 19.05\% in Val MAE / RMSE and 14.15\% / 5.21\% in Test MAE / RMSE. Even when compared with the stronger Average baseline, it still achieves additional improvements of 3.34\% / 6.54\% on Val and 2.85\% / 0.18\% on Test. A similar trend is observed for CLIP-Count, where the improvements over Mixed are even larger, reaching 37.55\% / 32.27\% on Val and 15.98\% / 1.81\% on Test. Compared with Average, our method further brings 16.47\% / 9.68\% gains on Val and 6.00\% / 0.49\% gains on Test. These results indicate that our approach is particularly effective for CLIP-based counting models under corrupted test conditions.

For CountGD, we also observe consistent improvements over both Mixed and Average. Specifically, compared with Mixed, our method achieves 22.27\% / 20.90\% error reduction on Val and 12.23\% / 1.08\% on Test; compared with Average, it further improves by 14.73\% / 15.14\% on Val and 7.87\% / 0.69\% on Test. Notably, these CountGD results are obtained \emph{without Adaptive Crop}. Therefore, the gains on CountGD should be regarded as conservative. In other words, even without the additional benefit of Adaptive Crop, our method still provides stable improvements, suggesting that the proposed adaptation strategy is effective and generalizes beyond CLIP-based architectures.

\begin{table}[htbp]
\centering
\caption{Performance under rain and fog corruption in outdoor scenes. Lower is better.}
\label{tab:outdoor_scene}
\setlength{\tabcolsep}{5pt}
\renewcommand{\arraystretch}{1.15}
\resizebox{0.8\linewidth}{!}{%
\begin{tabular}{llcccc}
\toprule
\textbf{Noise} & \textbf{TL-Denoiser} & \textbf{Val MAE} & \textbf{Val RMSE} & \textbf{Test MAE} & \textbf{Test RMSE} \\
\midrule

\multicolumn{6}{c}{\textbf{CounTX}} \\
\midrule
\cellcolor{cleanrow} Clean & \cellcolor{cleanrow} $\times$ 
& \cellcolor{cleanrow} 11.83 & \cellcolor{cleanrow} 58.44 
& \cellcolor{cleanrow} 14.88 & \cellcolor{cleanrow} 35.49 \\

\cellcolor{rainrow} Rain & \cellcolor{rainrow} $\times$ 
& \cellcolor{rainrow} 14.62 & \cellcolor{rainrow} 61.15 
& \cellcolor{rainrow} 17.98 & \cellcolor{rainrow} 41.59 \\

\cellcolor{tttrow} Rain & \cellcolor{tttrow} $\checkmark$ 
& \cellcolor{tttrow} \textbf{14.42} & \cellcolor{tttrow} \textbf{60.78} 
& \cellcolor{tttrow} \textbf{17.27} & \cellcolor{tttrow} \textbf{41.20} \\

\cellcolor{fogrow} Fog & \cellcolor{fogrow} $\times$ 
& \cellcolor{fogrow} 13.99 & \cellcolor{fogrow} 57.78 
& \cellcolor{fogrow} 15.76 & \cellcolor{fogrow} 34.57 \\

\cellcolor{tttrow} Fog & \cellcolor{tttrow} $\checkmark$ 
& \cellcolor{tttrow} \textbf{13.34} & \cellcolor{tttrow} \textbf{56.28} 
& \cellcolor{tttrow} \textbf{14.55} & \cellcolor{tttrow} \textbf{33.51} \\
\midrule

\multicolumn{6}{c}{\textbf{CLIP-Count}} \\
\midrule
\cellcolor{cleanrow} Clean & \cellcolor{cleanrow} $\times$ 
& \cellcolor{cleanrow} 16.80 & \cellcolor{cleanrow} 65.80 
& \cellcolor{cleanrow} 18.39 & \cellcolor{cleanrow} 39.21 \\

\cellcolor{rainrow} Rain & \cellcolor{rainrow} $\times$ 
& \cellcolor{rainrow} 35.11 & \cellcolor{rainrow} 104.14 
& \cellcolor{rainrow} 26.07 & \cellcolor{rainrow} 51.48 \\

\cellcolor{tttrow} Rain & \cellcolor{tttrow} $\checkmark$ 
& \cellcolor{tttrow} \textbf{21.64} & \cellcolor{tttrow} \textbf{92.42} 
& \cellcolor{tttrow} \textbf{19.74} & \cellcolor{tttrow} \textbf{38.10} \\

\cellcolor{fogrow} Fog & \cellcolor{fogrow} $\times$ 
& \cellcolor{fogrow} 18.77 & \cellcolor{fogrow} 69.67 
& \cellcolor{fogrow} 18.27 & \cellcolor{fogrow} 35.68 \\

\cellcolor{tttrow} Fog & \cellcolor{tttrow} $\checkmark$ 
& \cellcolor{tttrow} \textbf{18.47} & \cellcolor{tttrow} \textbf{69.10} 
& \cellcolor{tttrow} \textbf{17.93} & \cellcolor{tttrow} \textbf{35.42} \\
\midrule

\multicolumn{6}{c}{\textbf{CountGD}} \\
\midrule
\cellcolor{cleanrow} Clean & \cellcolor{cleanrow} $\times$ 
& \cellcolor{cleanrow} 9.49 & \cellcolor{cleanrow} 63.38 
& \cellcolor{cleanrow} 8.59 & \cellcolor{cleanrow} 21.40 \\

\cellcolor{rainrow} Rain & \cellcolor{rainrow} $\times$ 
& \cellcolor{rainrow} 9.86 & \cellcolor{rainrow} 65.52 
& \cellcolor{rainrow} 9.13 & \cellcolor{rainrow} 24.06 \\

\cellcolor{tttrow} Rain & \cellcolor{tttrow} $\checkmark$ 
& \cellcolor{tttrow} \textbf{9.52} & \cellcolor{tttrow} \textbf{63.56} 
& \cellcolor{tttrow} \textbf{8.69} & \cellcolor{tttrow} \textbf{22.38} \\

\cellcolor{fogrow} Fog & \cellcolor{fogrow} $\times$ 
& \cellcolor{fogrow} 11.54 & \cellcolor{fogrow} 68.96 
& \cellcolor{fogrow} 12.85 & \cellcolor{fogrow} 36.15 \\

\cellcolor{tttrow} Fog & \cellcolor{tttrow} $\checkmark$ 
& \cellcolor{tttrow} \textbf{9.74} & \cellcolor{tttrow} \textbf{63.36} 
& \cellcolor{tttrow} \textbf{8.84} & \cellcolor{tttrow} \textbf{22.34} \\
\bottomrule
\end{tabular}%
}
\end{table}
\paragraph{\textbf{Results under rain and fog}}
As shown in Table~\ref{tab:outdoor_scene}, different backbones exhibit different sensitivities to weather corruption. 
CLIP-Count is most affected by rain, whereas CountGD suffers more under fog, while CounTX shows relatively moderate degradation in both cases. 
Our method consistently improves all three backbones under both corruptions, with the largest gains observed in the most challenging settings. 
For example, it brings CLIP-Count under rain and CountGD under fog close to their clean-performance levels. 
These results demonstrate the robustness and cross-architecture effectiveness of the proposed method.

\begin{table}[htbp]
\centering
\caption{Inference time and parameter comparison of different methods} \label{tab:effiency}
\resizebox{0.45\textwidth}{!}{%
\begin{tabular}{lcc}
\hline
Methods & Inference Time (s) & Parameters (M) \\
\hline
CounTX            & 0.0312 & 161.08 \\
CounTX+Ours       & 0.0338 & 163.29 \\ \hline
CLIP-Count        & 0.0553 & 166.09 \\
CLIP-Count+Ours   & 0.0591 & 168.19 \\ \hline
CountGD            & 1.2174 & 233.36 \\
CountGD+Ours      & 1.2796 & 239.66 \\
\hline
\end{tabular}%
}
\end{table}

\paragraph{Efficiency analysis.}
As shown in Table~\ref{tab:effiency}, our method introduces only minor overhead in both inference time and parameter size. 
Across all three backbones, the inference-time increase is below 10\%, and the parameter growth is small relative to the original model scale. 
For example, on CounTX and CLIP-Count, our method increases the inference time by only 0.0026s and 0.0038s, respectively, with around 2M additional parameters. 
Even for the larger CountGD backbone, the extra cost remains modest. 
These results demonstrate that the proposed method is lightweight and practical, while achieving clear robustness gains under corrupted settings.

\begin{table}[htbp]
\centering
\caption{Comparison with different test-time training methods. Lower is better.}
\label{tab:com_ttt}
\setlength{\tabcolsep}{5pt}
\renewcommand{\arraystretch}{1.15}
\resizebox{0.82\linewidth}{!}{%
\begin{tabular}{llcccc}
\toprule
\textbf{Type} & \textbf{Method} & \textbf{Val MAE} & \textbf{Val RMSE} & \textbf{Test MAE} & \textbf{Test RMSE} \\
\midrule

\cellcolor{avgrow} Average & \cellcolor{avgrow} Source 
& \cellcolor{avgrow} 24.90 & \cellcolor{avgrow} 84.17 
& \cellcolor{avgrow} 19.49 & \cellcolor{avgrow} 109.77 \\

\cellcolor{mixedrow} Mixed & \cellcolor{mixedrow} Source 
& \cellcolor{mixedrow} 45.35 & \cellcolor{mixedrow} 139.29 
& \cellcolor{mixedrow} 25.85 & \cellcolor{mixedrow} 112.66 \\
\midrule

Average & Tent \cite{tent} 
& 24.84 & 84.05 
& 19.47 & 109.77 \\
Mixed & Tent \cite{tent} 
& 45.15 & 138.73 
& 25.79 & 112.59 \\

Average & EATA \cite{ETTA} 
& 24.53 & 83.56 
& 19.49 & 109.87 \\
Mixed & EATA \cite{ETTA} 
& 45.21 & 139.12 
& 25.53 & 111.83 \\

Average & CoTTA \cite{cotta} 
& 24.84 & 84.06 
& 19.47 & 109.77 \\
Mixed & CoTTA \cite{cotta} 
& 45.15 & 138.73 
& 25.79 & 112.58 \\

Average & TTT++ \cite{tttp} 
& 24.84 & 84.05 
& 19.47 & 109.77 \\
Mixed & TTT++ \cite{tttp} 
& 45.15 & 138.75 
& 25.79 & 112.60 \\
\midrule

\cellcolor{tttrow} Average & \cellcolor{tttrow} \textbf{Ours} 
& \cellcolor{tttrow} \best{20.80} & \cellcolor{tttrow} \best{76.02} 
& \cellcolor{tttrow} \best{18.32} & \cellcolor{tttrow} \best{109.23} \\

\cellcolor{tttrow} Mixed & \cellcolor{tttrow} \textbf{Ours} 
& \cellcolor{tttrow} \best{28.32} & \cellcolor{tttrow} \best{94.34} 
& \cellcolor{tttrow} \best{21.72} & \cellcolor{tttrow} \best{110.62} \\
\bottomrule
\end{tabular}%
}
\end{table}

\begin{figure*}[t]
    \centering
    \includegraphics[width=0.8\linewidth]{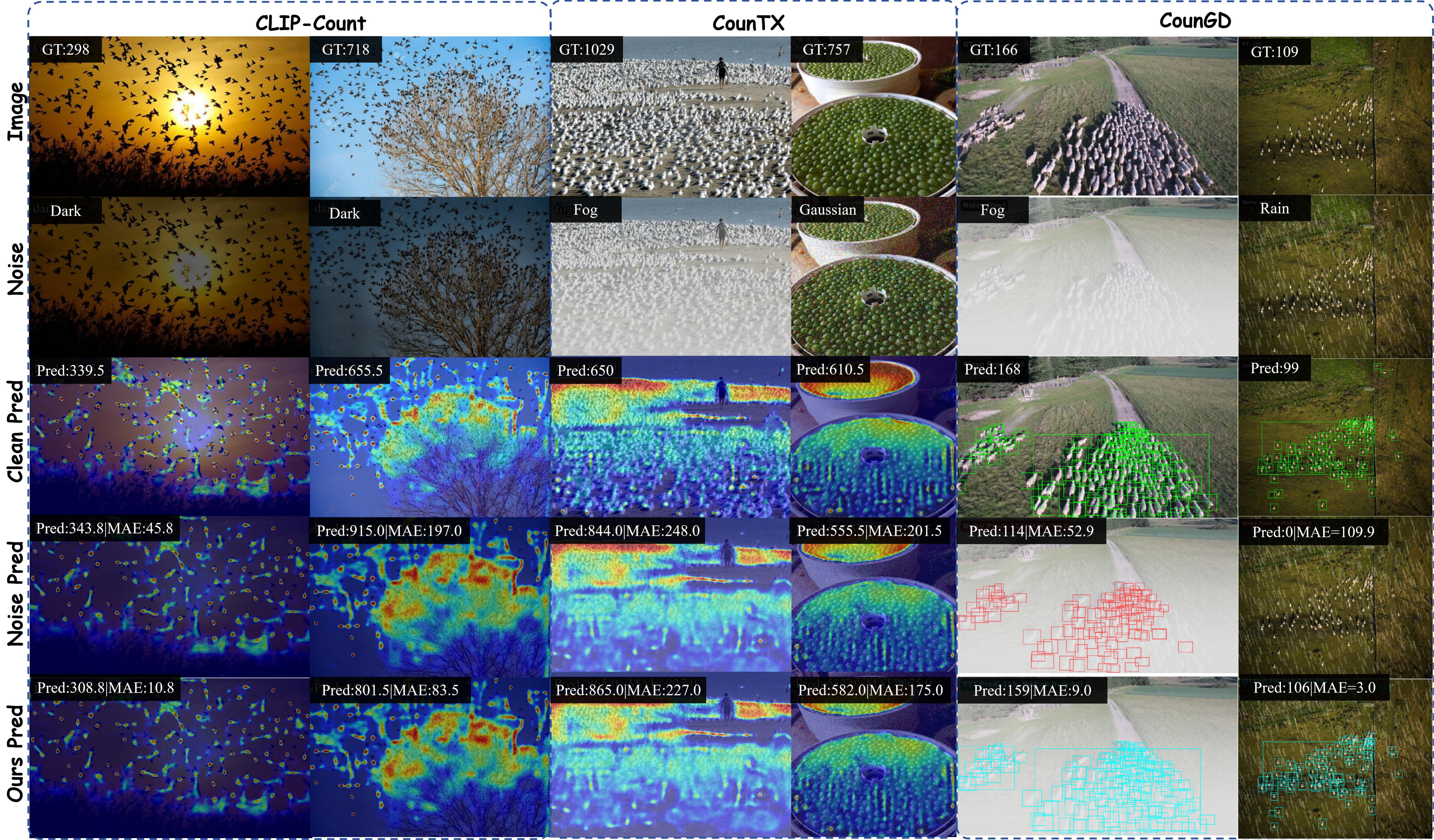}
    \caption{Qualitative comparison of different counting paradigms under corruptions.}
    \label{fig:res_vis}
\end{figure*}

\subsection{Comparison with TTA/TTT Baselines}
\paragraph{Comparison with existing test-time training methods.}
Table~\ref{tab:com_ttt} compares our method with several representative test-time adaptation/training baselines built on top of CLIP-Count. A clear observation is that existing generic methods, including Tent, EATA, CoTTA, and TTT++, bring only marginal improvements over the source model under both \emph{Average} and \emph{Mixed} settings. For example, under the \emph{Average} setting, Tent, CoTTA, and TTT++ only reduce the test MAE from 19.49 to 19.47, corresponding to a negligible 0.10\% improvement, while EATA yields no gain at all. A similar trend is observed under the more challenging \emph{Mixed} setting, where even the strongest competing baseline only reduces the test MAE from 25.85 to 25.53, i.e., by 1.24\%. These results suggest that existing generic test-time adaptation strategies are not well suited for open-vocabulary counting.

In contrast, our method achieves substantial gains under both settings. Under \emph{Average}, our method reduces the validation MAE/RMSE from 24.90/84.17 to 20.80/76.02, corresponding to relative improvements of 16.47\% and 9.68\%, respectively. On the test set, the MAE is reduced from 19.49 to 18.32, yielding a 6.00\% improvement. The gain becomes much more significant under \emph{Mixed}, where our method lowers the validation MAE from 45.35 to 28.32 and the validation RMSE from 139.29 to 94.34, corresponding to relative improvements of 37.55\% and 32.27\%, respectively. On the test set, the MAE decreases from 25.85 to 21.72, yielding a 15.98\% improvement. Compared with the strongest competing baseline, our method still achieves clear margins, reducing the test MAE by 5.91\% under \emph{Average} and 14.92\% under \emph{Mixed}.

Another notable observation is that the advantage of our method becomes more pronounced as the corruption setting becomes more challenging. While prior methods remain nearly ineffective under \emph{Mixed}, our method delivers substantial improvements in both MAE and RMSE, demonstrating strong robustness under severe distribution shifts. Overall, these results verify the necessity of designing a task-specific test-time training strategy for open-vocabulary counting, rather than directly adopting generic adaptation methods originally developed for classification.
\begin{table}[htbp]
\centering
\caption{Results under different module combinations. Lower is better.}
\label{tab:abl_study}
\setlength{\tabcolsep}{5pt}
\renewcommand{\arraystretch}{1.15}
\resizebox{0.95\linewidth}{!}{%
\begin{tabular}{cccllcccc}
\toprule
\textbf{Add} & \textbf{Mult} & \textbf{Entropy Gate} & \textbf{Setting} & \textbf{Metric} & \textbf{Val MAE} & \textbf{Val RMSE} & \textbf{Test MAE} & \textbf{Test RMSE} \\
\midrule

\cellcolor{baserow}  & \cellcolor{baserow}  & \cellcolor{baserow}  & \cellcolor{baserow} Baseline & \cellcolor{baserow} Average & \cellcolor{baserow} 21.24 & \cellcolor{baserow} 82.31 & \cellcolor{baserow} 17.22 & \cellcolor{baserow} 108.35 \\
\cellcolor{baserow}  & \cellcolor{baserow}  & \cellcolor{baserow}  & \cellcolor{baserow} Baseline & \cellcolor{baserow} Mixed   & \cellcolor{baserow} 42.34 & \cellcolor{baserow} 126.28 & \cellcolor{baserow} 25.30 & \cellcolor{baserow} 117.06 \\

\cellcolor{singlerow} \cmark & \cellcolor{singlerow}  & \cellcolor{singlerow}  & \cellcolor{singlerow} Add only & \cellcolor{singlerow} Average & \cellcolor{singlerow} 22.44 & \cellcolor{singlerow} 81.45 & \cellcolor{singlerow} 18.14 & \cellcolor{singlerow} 109.58 \\
\cellcolor{singlerow} \cmark & \cellcolor{singlerow}  & \cellcolor{singlerow}  & \cellcolor{singlerow} Add only & \cellcolor{singlerow} Mixed   & \cellcolor{singlerow} 32.43 & \cellcolor{singlerow} 96.60 & \cellcolor{singlerow} 21.92 & \cellcolor{singlerow} 110.46 \\

\cellcolor{singlerow}  & \cellcolor{singlerow} \cmark & \cellcolor{singlerow}  & \cellcolor{singlerow} Mult only & \cellcolor{singlerow} Average & \cellcolor{singlerow} 21.34 & \cellcolor{singlerow} 82.87 & \cellcolor{singlerow} 17.50 & \cellcolor{singlerow} 108.80 \\
\cellcolor{singlerow}  & \cellcolor{singlerow} \cmark & \cellcolor{singlerow}  & \cellcolor{singlerow} Mult only & \cellcolor{singlerow} Mixed   & \cellcolor{singlerow} 42.89 & \cellcolor{singlerow} 128.96 & \cellcolor{singlerow} 25.99 & \cellcolor{singlerow} 117.46 \\

\cellcolor{doublerow} \cmark & \cellcolor{doublerow} \cmark & \cellcolor{doublerow}  & \cellcolor{doublerow} Add+Mult & \cellcolor{doublerow} Average & \cellcolor{doublerow} 21.63 & \cellcolor{doublerow} 80.13 & \cellcolor{doublerow} 17.33 & \cellcolor{doublerow} 109.45 \\
\cellcolor{doublerow} \cmark & \cellcolor{doublerow} \cmark & \cellcolor{doublerow}  & \cellcolor{doublerow} Add+Mult & \cellcolor{doublerow} Mixed   & \cellcolor{doublerow} \best{32.01} & \cellcolor{doublerow} \best{96.23} & \cellcolor{doublerow} \best{21.06} & \cellcolor{doublerow} \best{110.30} \\

\cellcolor{fullrow} \cmark & \cellcolor{fullrow} \cmark & \cellcolor{fullrow} \cmark & \cellcolor{fullrow} Full model & \cellcolor{fullrow} Average & \cellcolor{fullrow} \best{20.53} & \cellcolor{fullrow} \best{76.93} & \cellcolor{fullrow} \best{16.73} & \cellcolor{fullrow} \best{108.06} \\
\cellcolor{fullrow} \cmark & \cellcolor{fullrow} \cmark & \cellcolor{fullrow} \cmark & \cellcolor{fullrow} Full model & \cellcolor{fullrow} Mixed   & \cellcolor{fullrow} 34.50 & \cellcolor{fullrow} 102.23 & \cellcolor{fullrow} 21.72 & \cellcolor{fullrow} 110.96 \\
\bottomrule
\end{tabular}%
}
\end{table}
\subsection{Ablation Studies}
Table~\ref{tab:abl_study} reports the ablation results of different module combinations on CounTX. 
The result shows that the additive branch (\textit{Add}) contributes the major performance gain, especially under the more challenging \emph{Mixed} setting, while the multiplicative branch (\textit{Mult}) mainly serves as a complementary refinement when combined with \textit{Add}. Further introducing the entropy gate improves stability and yields the best overall results under the \emph{Average} setting. Overall, the three components play different but complementary roles: \textit{Add} provides the main robustness gain, \textit{Mult} refines the adaptation, and the entropy gate enhances stability under milder shifts.
\subsection{Qualitative Analysis}
\paragraph{\textbf{Visualization of different models with TL-Denoiser}}
The Fig.~\ref{fig:res_vis} presents visual results of two density-estimation-based methods, CLIP-Count and CounTX, and one detection-based method, CountGD, under various corruptions such as Dark, Fog, Gaussian, and Rain. For each example, the first row shows the clean image and ground-truth count, and the second row shows the corrupted input. For CLIP-Count and CounTX, the subsequent rows visualize predicted density/response maps, where better methods produce spatial responses more aligned with object-populated regions and prediction counts closer to the ground truth. For CountGD, the results are shown as detection boxes, where improved performance is reflected by more accurate localization, fewer false positives/negatives, and smaller counting errors. Overall, the figure shows that our method improves robustness for both density-estimation and detection-based counting models under challenging corruptions.

\begin{figure}[h]
    \centering
    \includegraphics[width=0.85\linewidth]{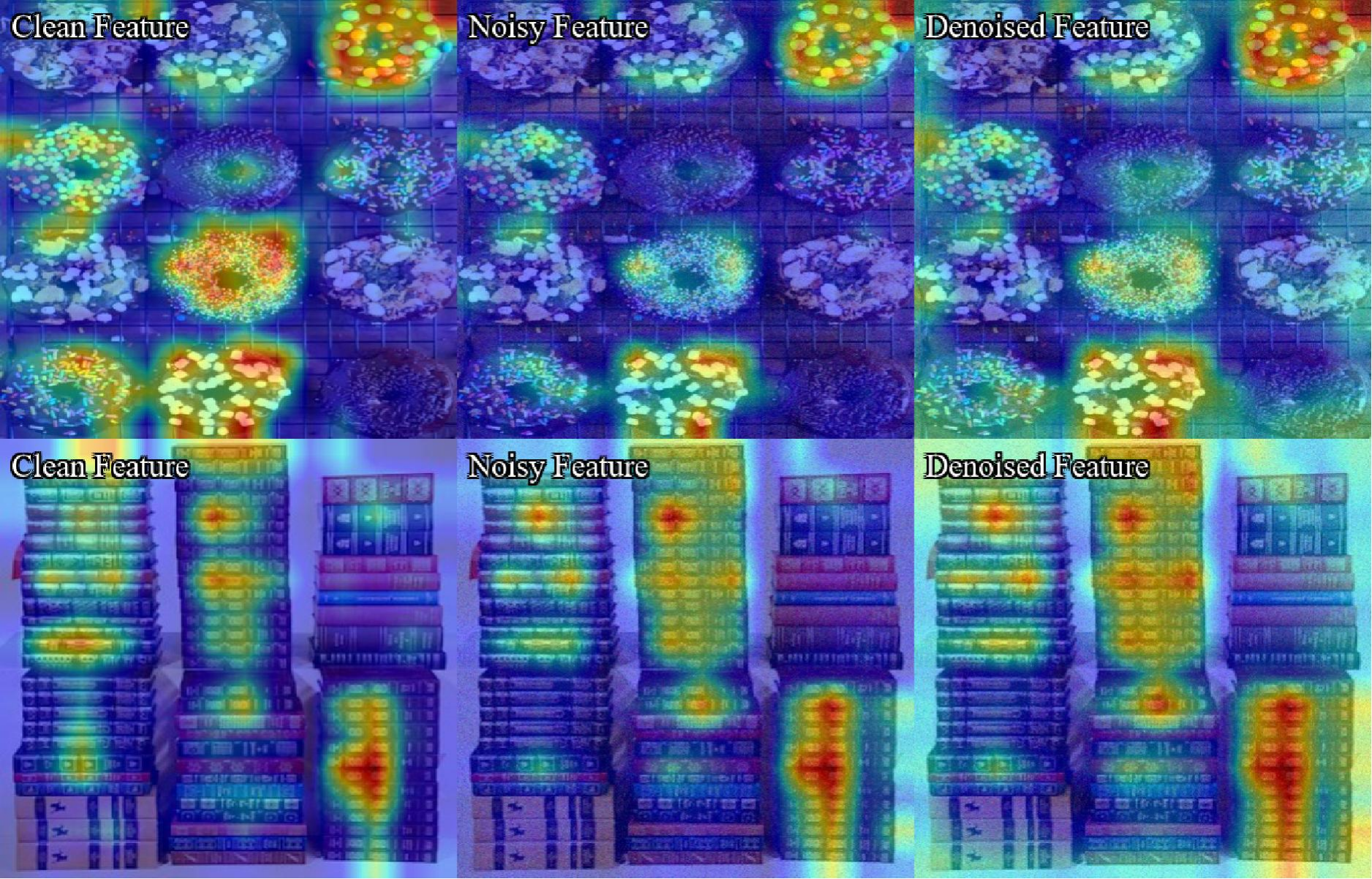}
    \caption{Feature visualization of clean, noisy, and denoised representations.}
    \label{fig:heatmap_vis}
\end{figure}

\paragraph{\textbf{Feature visualization of TL-Denoiser.}}
Fig.~\ref{fig:heatmap_vis} compares the feature responses of clean inputs, corrupted inputs, and denoised features. In the clean setting, the feature maps exhibit clear semantic structures, where the activated regions are well aligned with the underlying object distributions. After corruption is introduced, the noisy features become less coherent: the activations are more diffuse, local structures are blurred, and the boundaries between foreground objects and background regions become less distinguishable. This suggests that input corruption not only perturbs pixel appearance but also propagates into the feature space, leading to degraded semantic representations.
\section{Conclusion}

This paper takes the first step toward robust text-guided open-vocabulary object counting in real-world scenarios. 
We present \textbf{Robust-TOOC}, the first benchmark for systematically evaluating TOOC models under diverse corruptions, and propose \textbf{Dual-TTT}, the first test-time training framework for TOOC. 
By introducing a text-guided lightweight denoising module inspired by diffusion models, Dual-TTT can effectively mitigate corruption-induced feature degradation at inference time without requiring annotations or changing the original backbone architecture. 
Extensive experiments on multiple representative TOOC models verify the effectiveness, generality, and efficiency of our method in various types of corruption and severity levels. 
We hope that our benchmark and method will facilitate future research on robust and practical open-vocabulary counting systems.

\bibliographystyle{ACM-Reference-Format}
\bibliography{acmart}

\end{document}